\documentclass{article} 
\usepackage{nips15submit_e,times}
\usepackage{hyperref}
\hypersetup{pdfborder=0 0 0,
	colorlinks,
	citecolor=blue,
	linkcolor=blue,
	urlcolor=blue,
	}

\usepackage{graphicx} 
\usepackage{subfigure} 
\usepackage{amsthm}
\usepackage{amsmath}
\usepackage{amssymb}
\usepackage[numbers]{natbib}
\usepackage{tikz}
\usepackage{pgfplots}
\usepackage{ifpdf}
\usepackage{multirow}
\usepackage{booktabs}
\usepackage{xspace}
\usepackage{mathtools}
\usepackage{fancyref}
\usepackage{tabularx}

\usetikzlibrary{patterns}

\makeatletter
\newtheorem*{rep@theorem}{\rep@title}
\newcommand{\newreptheorem}[2]{%
\newenvironment{rep#1}[1]{%
 \def\rep@title{#2 \ref*{##1}\label{#1-rep}}%
 \begin{rep@theorem}}%
 {\end{rep@theorem}}}
\makeatother

\newtheorem{lemma}{Lemma}
\newtheorem{corollary}{Corollary}
\newtheorem{theorem}{Theorem}

\newreptheorem{lemma}{Lemma}
\newreptheorem{corollary}{Corollary}

\newcommand{\mailto}[1]{\href{mailto:#1}{#1}}

\DeclareMathOperator{\probability}{Pr}

\DeclareMathOperator{\CDF}{F}

\DeclareMathOperator{\rank}{rank}
\DeclareMathOperator{\TPR}{TPR}
\DeclareMathOperator{\FPR}{FPR}
\DeclareMathOperator{\PREC}{PRE}
\DeclareMathOperator{\FP}{FP}
\DeclareMathOperator{\TP}{TP}
\DeclareMathOperator{\FN}{FN}
\DeclareMathOperator{\TN}{TN}

\DeclareMathOperator*{\argmin}{arg\,min}
\DeclareMathOperator*{\argmax}{arg\,max}
\DeclareMathOperator*{\subjectto}{subject\,to}
\DeclareMathOperator*{\topfun}{head}
\DeclareMathOperator*{\bottomfun}{tail}
\newcommand{\floor}[1]{\big\lfloor #1 \big\rfloor}
\newcommand{\ceil}[1]{\big\lceil #1 \big\rceil}

\newcommand{\pfrac}{\beta\xspace}

\newcommand{\posprobability}{\mathcal{P}}
\newcommand{\knownpos}{\posprobability_L}
\newcommand{\knownneg}{\mathcal{N}_L}
\newcommand{\latentpos}{\posprobability_U}
\newcommand{\surrpos}{\posprobability_U^\star}
\newcommand{\surrposopt}{\posprobability_U^\ast}
\newcommand{\surrposbullet}{\posprobability_U^\bullet}
\newcommand{\bothpos}[1][]{\posprobability^{#1}_\Omega}
\newcommand{\bothposapprox}[1][]{\posprobability^{\star}_{\Omega}}
\newcommand{\bothposbullet}[1][]{\posprobability^{\bullet}_{\Omega}}
\newcommand{\overall}{\mathcal{R}}

\newcommand{\cdf}[2]{\CDF(#1, #2)}

\newcommand{\posset}{\mathcal{P}}
\newcommand{\pos}{\posset}
\newcommand{\unlabeled}{\mathcal{U}}
\newcommand{\tprsymb}{t}
\newcommand{\ranksymb}{r}

\title{Assessing Binary Classifiers Using Only Positive and Unlabeled Data}

\author{
Marc Claesen \\
Dept. of Electrical Engineering, STADIUS \\
KU Leuven \& iMinds Medical IT\\
\mailto{marc.claesen@esat.kuleuven.be} 
\And
Jesse Davis \\
Dept. of Computer Science, DTAI \\
KU Leuven \\
\mailto{jesse.davis@cs.kuleuven.be} 
\And
Frank De Smet \\
Dept. of Public Health and Primary Care \\
KU Leuven \\
\And
Bart De Moor \\
Dept. of Electrical Engineering, STADIUS \\
KU Leuven \& iMinds Medical IT \\
}

\nipsfinalcopy 

\begin{document}

\maketitle

\begin{abstract}
Assessing the performance of a learned model is a crucial part of machine learning. However, in some domains only positive and unlabeled examples are available, which prohibits the use of most standard evaluation metrics. We propose an approach to estimate any metric based on contingency tables, including ROC and PR curves, using only positive and unlabeled data. Estimating these performance metrics is essentially reduced to estimating the fraction of (latent) positives in the unlabeled set, assuming known positives are a random sample of all positives. We provide theoretical bounds on the quality of our estimates, illustrate the importance of estimating the fraction of positives in the unlabeled set and demonstrate empirically that we are able to reliably estimate ROC and PR curves on real data. 
\end{abstract}


\section{Introduction} \label{introduction}

Model evaluation is a critical step in the learning process. Typically, evaluations either report summary metrics, such as accuracy, F1 score, or area under the receiver operator characteristic (ROC) curve or visually show a model's performance under different operating conditions by using ROC or precision-recall curves. All the aforementioned evaluation approaches require constructing contingency tables (also called confusion matrices), which show how a model's predicted labels relate to an example's ground truth label. Computing a contingency table requires labeled examples. However, for many problems only a few labeled examples and many unlabeled ones are available as acquiring labels can be time-consuming, costly, unreliable, and in some cases impossible.


The field of semi-supervised learning \citep{chapelle2006semi} focuses on coping with partially labeled data. Positive and unlabeled (PU) learning is a special case of semi-supervised learning where each example's label is either positive or not known~\citep{liu2003building, yu2004pebl, denis2005learning, Elkan:2008:LCO:1401890.1401920, scottblanchard,mordelet2014bagging, claesen2014robust}. Both semi-supervised and PU learning tend to focus on developing learning algorithms that cope with partially labeled data during training as opposed to evaluating algorithms when the test set is partially labeled. What is less well studied is the effect of partially labeled data on evaluation. Currently, algorithms are evaluated assuming that the test data is fully labeled \citep{goldman2000enhancing, nigam2000text, chawla2005learning, calvo2007learning, chapelle2008optimization, mordelet2014bagging, claesen2014robust} and if the test data is only partially labeled, sometimes it is assumed that all unlabeled instances are negative when evaluating performance~\citep{mordelet2011prodige, sifrim2013extasy,sechidis2014statistical}.

This paper describes how to incorporate the unlabeled data in the model evaluation process. We show how to compute contingency tables based on only positive and unlabeled examples where the unlabeled set contains both positive and negative examples, by looking at the ranking of examples produced by a model.
Theoretically, we establish important relationships between contingency tables and rank distributions, which allow us to provide bounds on the false positive rate at each rank when the ranking contains examples whose ground truth label is unknown. Our findings have important implications for model selection as we show that naively assuming that all unlabeled examples are negative, as is sometimes done in PU learning, could lead to selecting the wrong model. We demonstrate the efficacy of our approach by estimating ROC and PR curves from real-world data.

\section{Background and definitions}
We first review the relevant background on model evaluation and issues caused by partial labeling. 



\subsection{Rank distributions and contingency tables} \label{contingency-intro}

We focus on binary decision problems, where the goal is to classify examples as either positive or negative. Most learned models (e.g., SVM, logistic regression, naive Bayes) predict a numeric score for each example where higher values imply higher confidence that the instance belongs to the positive class. Typically, a \emph{ranking} $\overall$ is produced by sorting examples in descending order by their numeric score such that confident positive predictions are ranked close to the top of $\overall$.\footnote{Which means a low value for rank in this work, though this is often referred to as \emph{highly ranked} in literature.}  




Within a ranking $\overall$, we treat $\pos \subset \overall$ as the subset of examples with positive labels, $\bar{\pos} = \overall - \pos$  as the subset of examples with negative labels, and let $\rank(\overall, x)$ denote the rank of an instance $x$ in $\overall$.  Given a cutoff rank $r,$ predictions can be made by assigning the positive class to the $r$ top ranked instances and the negative class to the rest. This decision rule yields a \emph{true positive rate (TPR)}, which is the fraction of positive examples that are correctly labeled as positive, and \emph{false positive rate (FPR)}, which is the fraction of negative examples that are incorrectly labeled as positive:
\begin{align}
\TPR(\pos, r) &= \probability(\rank(\overall, x) \leq r\ |\ x \in \pos) = |\{\ x \in \pos\ :\ \rank(\overall, x) \leq r\}|\ /\ |\pos|, \label{tpr-def} \\
\FPR(\pos, r) &= \probability(\rank(\overall, \bar{x}) \leq r\ |\ \bar{x} \in \bar{\pos}) = \TPR(\overall - \pos, r). \label{fpr-def}
\end{align}
\noindent Given the number of positives $|\pos|$ and negatives $|\overall-\pos|$, the contingency table for a rank $r$ is:
\begin{minipage}[c]{0.47\textwidth}
\begin{align}
\TP(\pos,r) &= \TPR(\pos, r) \cdot |\pos|, \label{tp-def} \\
\FN(\pos,r) &= |\pos| - \TP(\pos, r), 
\end{align}
\end{minipage}\hfill\begin{minipage}[c]{0.47\textwidth}
\begin{align}
\FP(\pos,r) &= \FPR(\pos, r) \cdot |\overall-\pos|, \nonumber \\
\TN(\pos,r) &= |\overall-\pos| - \FP(\pos,r). 
\end{align}
\end{minipage}

\begin{figure}[ht]
  \centering
  \begin{minipage}[c]{0.40\textwidth}
  \includegraphics[height=2.5cm]{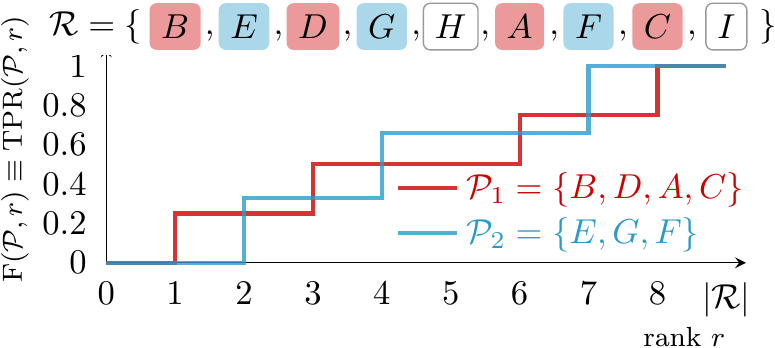}
  \end{minipage}\ \ 
  \begin{minipage}[c]{0.58\textwidth}
\addtolength{\abovecaptionskip}{-3mm}
  \caption{Rank CDF of two sets of positives $\pos_1 = \{B, D, A, C\}$ and $\pos_2=\{E, G, F\}$ within an overall ranking $\overall=\{B,E,D,G,H,A,F,C,I\}$, with $|\pos_1|=4$ and $|\pos_2|=3$. In practice $\overall$ is obtained by sorting the data according to classifier score. The rank CDF of a set $\mathcal{S}\subseteq \mathcal{R}$ is based on the positions of elements of $\mathcal{S}$ in $\overall$. } 
  \label{fig:rank-cdf}
\addtolength{\abovecaptionskip}{3mm}
  \end{minipage}
\end{figure}

The rank distribution of a set of instances $\pos$ within an overall ranking $\overall$ is defined as the distribution of their corresponding ranks within $\overall$. The rank cumulative distribution function (CDF) of a set of instances $\pos$ is defined as the (empirical) CDF of their ranks, i.e. $\forall\ r \in \{1,\ldots,|\overall|\}$:
\begin{equation}
\cdf{\pos}{r} = \probability(\rank(\overall, x) \leq r \ |\ x \in \pos). \label{rankcdf}
\end{equation}
The concept of rank CDF is illustrated in Figure~\ref{fig:rank-cdf}. Note that $ \cdf{\pos}{r} \equiv \TPR(\pos, r) $ (Equations~\eqref{tpr-def} and~\eqref{rankcdf}), that is, the rank CDF of the set of positives $\pos$ at rank $r$ in an overall ranking $\overall$ can be interpreted directly as a true positive rate, when labeling the $r$ top ranked instances as positive.
\def\floatdelta{5mm}
\addtolength{\textfloatsep}{-\floatdelta}
\addtolength{\textfloatsep}{\floatdelta}

We use two convenience functions to partition sets of ranks:
\begin{equation*}
\topfun(X, \ranksymb) = \{\ \rank(\overall, x) \leq \ranksymb\ :\ x \in X\ \} \text{ and }
\bottomfun(X, \ranksymb) = \{\ \rank(\overall, x) > \ranksymb\ :\ x \in X\ \},
\end{equation*}
such that $\topfun(X,\ranksymb)\ \cup\ \bottomfun(X,\ranksymb) = X$ and $|\topfun(X,r)| = \cdf{X}{\ranksymb}\cdot |X|$. 


\subsection{ROC and PR curves} \label{roc}
Receiver operator characteristic (ROC) curves are used extensively for evaluating classifiers in machine learning \citep{Bradley:1997:UAU:1746432.1746434} as they illustrate the performance of a model over its entire operating range. ROC curves depict how a model's true positive rate (shown on the y-axis) varies as a function of its false positive rate (shown on the x-axis). Each cutoff rank $r \in \{1,\ldots,|\overall|\}$ corresponds to a single point (i.e., (FPR, TPR) pair) in ROC space (Eqs.~\eqref{tpr-def} and \eqref{fpr-def}). An (empirical) ROC curve for a ranking $\overall$ and set of positives $\pos \subset \overall$ is constructed by computing $\FPR(\pos, r)$ and $\TPR(\pos, r)$ at each rank $r$ and interpolating by drawing a straight line between points corresponding to consecutive ranks.  
The area under an ROC curve (AUROC) is a commonly used summary statistic, typically ranging between $0.5$ (random model) and $1$ (perfect model). AUROC is a popular  criterion in model selection and is often used as the optimization objective in hyperparameter search \citep{Bradley:1997:UAU:1746432.1746434}.



Precision-Recall (PR) curves  \citep{davis2006relationship} are an alternative to ROC curves that show how a model's precision (y-axis) varies as a function of recall (x-axis). Recall is equivalent to TPR and precision is the fraction of examples classified as
positive that are truly positive ($\TP / (\TP+\FP)$). PR curves are widely used when there is a skew in the class distributions~\cite{davis-icml09,claesen2014robust}.




\subsection{Evaluation with partially labeled data}
In the partial labeling setting, $\overall$ consists of disjoint sets of known positives $\knownpos$, known negatives $\knownneg$ and unlabeled instances $\unlabeled$. 
The unlabeled set $\unlabeled$ consists of latent positives $\latentpos$ and latent negatives. The fraction of latent positives in the unlabeled set plays a crucial role in our work, denoted by $\beta$:
\begin{equation}
\pfrac = \probability(x \in \latentpos\ |\ x \in \unlabeled) = |\latentpos|\ /\ |\unlabeled|. \label{pfrac} 
\end{equation}

Note that computing contingency tables requires fully labeled data. If only a few labeled instances of both classes are available, they can be used to compute rough estimates of predictive performance. However, if only positive labels are available, even a rough approximation of common metrics cannot be estimated directly as we do not know which unlabeled examples are positives and which are negative. A common approach to evaluate models in a PU learning context is to treat the full unlabeled set as negative \citep{mordelet2011prodige,sifrim2013extasy,sechidis2014statistical}, though we will show that this may lead to spurious results.

\section{Relationship between the rank CDF of positives and contingency tables} \label{rank-roc}
The challenge of incorporating unlabeled data into an evaluation metric is knowing which unlabeled examples are latent positives and which are latent negatives. Our insight is that, if the known positives are sampled completely at random from all positives, the rank distribution of latent positives should follow the rank distribution of known positives. Thus if we know $\pfrac$, which is needed to compute the expected number of latent positives within the unlabeled data, this provides an avenue for building contingency tables that incorporate the unlabeled data. To do so, we first prove relationships between rank CDFs of sets of positives within an overall ranking at a given rank $r$ and the corresponding contingency tables. Then, we use these relationships to prove bounds on the FPR at a given rank $r$ when the ranking includes unlabeled examples, some of which are latent positives. 




\subsection{Rank distributions and contingency tables based on subsets of positives within a ranking}
We begin by considering given sets of positives within an overall ranking. Proofs of all lemmas can be found in Appendix~\ref*{proofs}, along with figures to illustrate the associated property.

\begin{replemma}{lemma-rank-fpr}
Given a rank $\ranksymb$ and two disjoint subsets of positives $\pos_1$ and $\pos_2$ within an overall ranking $\overall$. If $|\pos_1|=|\pos_2|$ and $\TPR(\pos_1, \ranksymb) > \TPR(\pos_2,\ranksymb)$, then $\FPR(\pos_1,\ranksymb) < \FPR(\pos_2,\ranksymb)$.
\end{replemma}


\begin{replemma}{lemma-rank-union}
Given a rank $\ranksymb$ and two disjoint sets of positives $\pos_1$, $\pos_2$ in a ranking $\overall$ and $\bothpos=\pos_1\cup\pos_2$. If $\TPR(\pos_1,\ranksymb) < \TPR(\pos_2, \ranksymb)$ then $\TPR(\pos_1,\ranksymb) < \TPR(\bothpos,\ranksymb) < \TPR(\pos_2, \ranksymb)$.
\end{replemma}

\begin{repcorollary}{corollary-rank-union}
Given a rank $\ranksymb$ and three sets of positives $\pos_A$, $\pos_B$ and $\pos_C$ within a ranking $\overall$ such that $\pos_A\cap\pos_B=\emptyset$ and $\pos_A\cap\pos_C=\emptyset$ and $|\pos_B|=|\pos_C|$, then
\begin{equation*}
\TPR(\pos_B,\ranksymb) < \TPR(\pos_C,\ranksymb) \quad\leftrightarrow\quad \TPR(\pos_A\cup \pos_B,\ranksymb) < \TPR(\pos_A\cup\pos_C,\ranksymb).
\end{equation*}
\end{repcorollary}

\ifx
\begin{figure}[!h]
  \centering
  \RawFloats
  \begin{minipage}[b]{0.49\textwidth}
  \includegraphics[width=\textwidth]{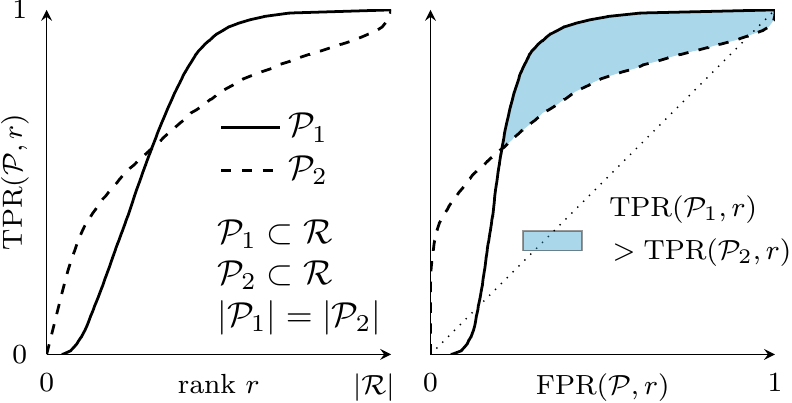}
  \caption{Illustration of Lemma~\ref{lemma-rank-fpr}: higher TPR at a given rank $r$ implies lower FPR at $r$ for two positive sets of the same size.}
  \label{fig:lemma-rank-fpr}
  \vfill
  \end{minipage}
  \hfill
  \begin{minipage}[b]{0.49\textwidth}
  \includegraphics[width=\textwidth]{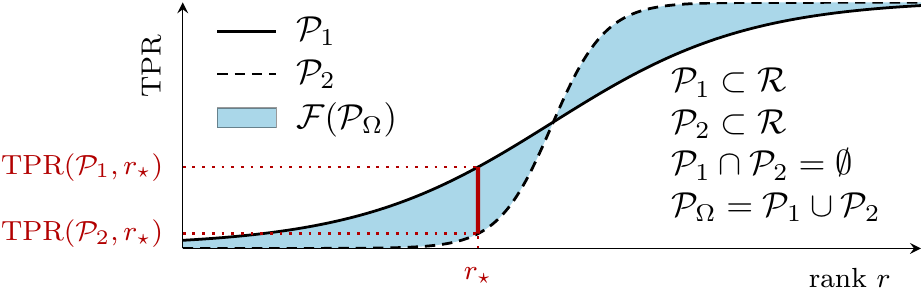}
  \caption{Illustration of Lemma~\ref{lemma-rank-union}: $\mathcal{F}(\cdot)$ denotes feasible region. The rank distribution of the union $\bothpos$ of two sets of positives $\pos_1$ and $\pos_2$ lies between their respective rank distributions.} 
  \label{fig:lemma-rank-union}
\vfill
  \end{minipage}
\end{figure}
\fi

\subsection{Contingency tables based on partially labeled data}
Lemmas~\ref*{lemma-rank-fpr} and \ref*{lemma-rank-union} describe relationships between rank distributions and contingency tables of different (but known) sets of positives within an overall ranking. 
We now show how to construct contingency tables corresponding to the greatest-lower and least-upper bound of the FPR at a given rank, accounting for the unknown set of latent positive example from partially labeled data, given $\pfrac$. 





\begin{theorem} \label{main-theorem}
Given an overall ranking $\overall$ consisting of disjoint sets of known positives $\knownpos$, known negatives $\knownneg$ and unlabeled instances $\unlabeled$, where $\unlabeled$ contains an unknown set of latent positives $\latentpos \subset \unlabeled$ of known size $|\latentpos| = \pfrac \cdot|\unlabeled|$. Given a rank $\ranksymb$ and an upper bound $\mathcal{T}_{ub}(\ranksymb) \geq \TPR(\latentpos, \ranksymb)$, a tight lower bound on $\FPR(\bothpos, \ranksymb)$ with $\bothpos = \knownpos \cup \latentpos$ can be found without explicitly identifying $\latentpos$.

\textbf{Proof}: Step 1: assign a set of surrogate positives $\surrpos$:\footnote{A surrogate positive is an example that we treat as if its ground truth label is positive (even though in reality its ground truth label is unknown) when constructing a contingency table.}
\begin{align}
\surrpos = \argmin_{\surrposopt\subset\ \unlabeled}\ &\TPR(\surrposopt, \ranksymb)\ \label{eq:surrpos-def} 
\subjectto\ \TPR(\surrposopt,\ranksymb) \geq \mathcal{T}_{ub}(\ranksymb) \text{ and } |\surrposopt|=\pfrac\cdot|\unlabeled|, 
\end{align}
then $\TPR(\surrpos,\ranksymb) \geq \TPR(\latentpos,\ranksymb)$ by construction. If $|\topfun(\unlabeled, r)| < \pfrac\mathcal{T}_{ub}(\ranksymb)\cdot |\unlabeled|$, then no $\surrpos$ exists that satisfies the constraint $\TPR(\surrpos, \ranksymb) \geq \mathcal{T}_{ub}(\ranksymb)$ in Equation~\eqref{eq:surrpos-def}.\footnote{An infeasibility implies that $\mathcal{T}_{ub}(\ranksymb)$ and/or $\pfrac$ are too high.} In this case, treat all instances in $\topfun(\unlabeled, \ranksymb)$ as surrogate positive, which trivially implies $\TPR(\surrpos, \ranksymb) \geq \TPR(\latentpos, \ranksymb)$.

Step 2: define $\bothposapprox = \knownpos \cup \surrpos$. Using Corollary~\ref{corollary-rank-union} yields $\TPR(\bothposapprox, \ranksymb) \geq \TPR(\bothpos, \ranksymb)$. Since $|\bothposapprox|=|\bothpos|$, using Lemma~\ref*{lemma-rank-fpr} yields the lower bound on FPR, i.e., $\FPR(\bothposapprox, \ranksymb) \leq \FPR(\bothpos, \ranksymb)$.\hfill$\blacksquare$
\end{theorem}

Applying Theorem~\ref{main-theorem} yields a nontrivial lower bound on $\FPR(\bothpos,\ranksymb)$. In Lemma~\ref*{lemma-glb} we prove that $\FPR(\bothposapprox,\ranksymb)$ is the greatest achievable lower bound based on a given $\unlabeled\subset\overall$.

\ifx
\begin{figure}[!h]
  \centering
  \includegraphics[width=\textwidth]{theorem-1-figure.pdf}
  \caption{Explanation} \label{fig:theorem-1}
\end{figure}
\fi

\begin{replemma}{lemma-glb}
Minimizing $\TPR(\surrpos, \ranksymb)$ in Equation~\eqref{eq:surrpos-def} of Theorem~\ref{main-theorem} ensures $\FPR(\bothposapprox, \ranksymb)$ is the greatest achievable lower bound on $\FPR(\bothpos, \ranksymb)$ given $\pfrac$, $\mathcal{T}_{ub}(\ranksymb)$, $\overall$ and $\unlabeled$. 
\end{replemma}

Due to its symmetry, Theorem~\ref{main-theorem} can also be used to obtain the least achievable upper bound of $\FPR(\bothpos, \ranksymb)$ given a ranking $\overall$ and a bound $\mathcal{T}_{lb}(\ranksymb) \leq \TPR(\surrpos, \ranksymb)$ by assigning $\surrpos$ such that:
\begin{align}
\surrpos = \argmax_{\surrposopt\subset\ \unlabeled}\ &\TPR(\surrposopt, \ranksymb)\ \label{eq:surrpos-def-ub} 
\subjectto\ \TPR(\surrposopt,\ranksymb) \leq \mathcal{T}_{lb}(\ranksymb) \text{ and } |\surrposopt|=\pfrac\cdot|\unlabeled|. 
\end{align}

\section{Efficiently computing the bounds} \label{practical}



We now describe how to use Theorem~\ref{main-theorem} and Lemma~\ref*{lemma-glb} to compute the contingency tables corresponding to the greatest lower and least upper bound on $\FPR(\bothpos,\ranksymb)$ from a finite sample. First, we explain how to compute contingency tables efficiently via Theorem~\ref{main-theorem}. Second, we propose how to obtain the bounds on rank CDF ($\mathcal{T}_{lb}(\ranksymb)$ and $\mathcal{T}_{ub}(\ranksymb)$) that are needed to build the contingency table.


\subsection{Computing the contingency table with greatest-lower bound on FPR at given rank $r$} \label{contingency}
Given $\pfrac$, $\overall$ and the sets $\knownpos$, $\knownneg$, and $\unlabeled$, Theorem~\ref{main-theorem} enables computing contingency tables corresponding to the least upper and greatest lower bound on FPR at a given cutoff rank $r$. We focus on building the contingency table corresponding to the lower bound on the FPR, the other is analogous. 


%


 We decompose the computation to consider the labeled and unlabeled instances separately: 
\begin{equation*}
\begin{bmatrix}
\TP_{\Omega}^r & \FP_{\Omega}^r \\
\FN_{\Omega}^r & \TN_{\Omega}^r
\end{bmatrix} = 
\begin{bmatrix*}[l]
\TP_{L}^r = |\topfun(\knownpos, r)| & \FP_{L}^r = |\topfun(\knownneg, r)|\\
\FN_{L}^r = |\bottomfun(\knownpos, r)| & \TN_{L}^r = |\bottomfun(\knownneg, r)|
\end{bmatrix*}
+
\begin{bmatrix}
\TP_{U}^r & \FP_{U}^r\\
\FN_{U}^r & \TN_{U}^r
\end{bmatrix}.
\end{equation*}
Given that at rank $r$ we can directly compute partial contingency tables for the labeled data based on $\overall$, $\knownpos$ and $\knownneg$, we focus on computing the contingency table for the unlabeled instances. 

Given $\mathcal{T}_{ub}(r)$, we can use Theorem~\ref{main-theorem} to determine the values in the contingency table for the unlabeled instances for the greatest lower bound on FPR. Doing so requires inferring a set of surrogate positives $\surrpos$ from the unlabeled data, which must be a solution to Equation~\eqref{eq:surrpos-def}. This requires $\theta$ surrogate positives in $\topfun(\surrpos,\ranksymb)$ and the rest in $\bottomfun(\surrpos,\ranksymb)$, where $\theta$ is defined as:
\begin{equation}
\theta = \ceil{\mathcal{T}_{ub}(r) \cdot |\surrpos|} = \ceil{\mathcal{T}_{ub}(r) \cdot \pfrac \cdot |\unlabeled|}, \label{theta}
\end{equation}
By rounding up in Equation~\eqref{theta}, we ensure that $\TPR(\surrpos, r) \geq \mathcal{T}_{ub}(r)$ as required by Theorem~\ref{main-theorem}. 

In practice, two corner cases must be considered. One is if $|\topfun(\unlabeled,\ranksymb)| < \theta$, then it is impossible to assign $\theta$ surrogates below rank $\ranksymb$ in $\unlabeled$. In this case, all of $\topfun(\unlabeled,\ranksymb)$ is assigned as surrogate positives and the remaining surrogates are in $\bottomfun(\unlabeled,\ranksymb)$ (as discussed in Theorem~\ref{main-theorem}). 
Two is if $|\bottomfun(\unlabeled, r)| < |\surrpos| - \theta$, in which case all of $\bottomfun(\unlabeled, r)$ is labeled positive and the remaining surrogate positives inevitably end up in $\topfun(\unlabeled, r)$. Hence, any set of surrogate positives $\surrpos$ that meets the following criteria solves Equation~\eqref{eq:surrpos-def} and thus yields a valid bound:
\begin{equation}
|\surrpos| = \pfrac \cdot |\unlabeled| \text{ and } |\topfun(\surrpos,r)| = 
\left\{\begin{matrix*}[l] 
\min\big(|\topfun(\unlabeled, r)|,\ \theta\big) &\text{if } |\surrpos|-\theta \leq |\bottomfun(\unlabeled, r)|, \\
|\surrpos| - |\bottomfun(\unlabeled, r)| &\text{if } |\surrpos|-\theta > |\bottomfun(\unlabeled, r)|.
\end{matrix*}\right. \label{eq:surrpos-conditions}
\end{equation}

Given a set of surrogate positives $\surrpos$, the partial contingency table of interest becomes:
\begin{equation}
\begin{bmatrix}
\TP_{U}^r & \FP_{U}^r\\
\FN_{U}^r & \TN_{U}^r
\end{bmatrix} = 
\begin{bmatrix*}[l]
|\topfun(\surrpos,\ranksymb)| 		& |\topfun(\unlabeled-\surrpos,\ranksymb)|\\
|\bottomfun(\surrpos,\ranksymb)| 	& |\bottomfun(\unlabeled-\surrpos,\ranksymb)|
\end{bmatrix*}, \label{eq:partial-ct-unlabeled}
\end{equation}
where $\unlabeled-\surrpos$ is the set of surrogate negatives and $|\surrpos|$ and $|\topfun(\surrpos,r)|$ are known via Eq.~\ref{eq:surrpos-conditions}. 


Note that computing the partial contingency table for the unlabeled data can be done very efficiently since it only requires set sizes as shown in Equation~\ref{eq:partial-ct-unlabeled}, without explicitly partitioning the unlabeled set $\unlabeled$. That is, we do not need to know which examples are in $\topfun(\surrpos,r)$, $\bottomfun(\surrpos,r)$, $\topfun(\unlabeled-\surrpos,r)$ and $\bottomfun(\unlabeled-\surrpos,r)$, we just need to know the number of examples each set contains.

The contingency table with least upper bound on $\FPR(\latentpos,\ranksymb)$ is obtained by replacing Eq.~\eqref{theta} by:
\begin{equation}
\theta = \floor{\mathcal{T}_{lb}(r) \cdot |\surrpos|} = \floor{\mathcal{T}_{lb}(r) \cdot \pfrac \cdot |\unlabeled|}. \label{theta-alt}
\end{equation}

\subsection{Bounds on the rank distribution of $\latentpos$} \label{quantile-bounds}
Applying Theorem~\ref{main-theorem} to build a contingency table at rank $\ranksymb$ requires a bound $\mathcal{T}_{ub}(\ranksymb) \geq \TPR(\latentpos, \ranksymb)$ for estimating a lower bound on the FPR and a bound $\mathcal{T}_{lb}(\ranksymb) \leq \TPR(\latentpos, \ranksymb)$  for estimating an upper bound on the FPR. To compute these bounds, we assume known and latent positives have similar rank distributions. This holds when known positives $\knownpos$ are selected completely at random from all positives $\bothpos$, but is violated if the process of selecting examples for labeling is biased \citep{chawla2005learning}. 

$\TPR(\bothpos,\ranksymb)$ is estimated via the empirical rank CDF of $\knownpos$, which only approximates the true CDF. To acccount for uncertainty, we construct confidence intervals (CIs) for the CDF. Our assumption implies that a CI of the CDF based on $\knownpos$ is also a CI of the CDF of $\latentpos$. A CI boundary is treated as a function mapping rank $r$ to the estimated bound on the CDF. $\mathcal{T}_{lb}$ and $\mathcal{T}_{ub}$ denote these bounds:
\begin{equation}
0 \leq \mathcal{T}_{lb}(r) \leq \TPR(\knownpos, \ranksymb), \TPR(\latentpos,\ranksymb), \TPR(\bothpos, \ranksymb) \leq \mathcal{T}_{ub}(r) \leq 1,\ \forall\ r. \label{eq:ci}
\end{equation}
We formalize the bounds of the CI of the CDF as functions of rank because an underlying set with that rank distribution does not necessarily exist in the overall ranking $\overall$.

The confidence band on rank CDF can be computed based on the known positives in several ways. We use a standard bootstrap approach \citep{efron1994introduction} in our experiments. 
Having many known positives yields a tight confidence band on rank CDF, which then translates to tight bounds on performance metrics.



\section{Constructing ROC and PR curve estimates} \label{roc-pr}

Next, we describe how to estimate bounds on the true ROC and PR curves. Though we focus on these two criteria, our approach can be used to estimate any metric based on contingency tables.

{\bf ROC curves}
Given a ranking, instead of constructing a single ROC curve, our approach computes two curves: one corresponding to the upper bound and one corresponding to the lower bound on the CI on rank CDF of known positives $\knownpos$, using the methodology outlined in Section 4 to compute two contingency tables for each rank $\ranksymb$, corresponding to the greatest lower and least upper bound on $\FPR(\bothpos,r)$. The set of contingency tables corresponding to greatest lower bounds on FPR at each rank form an upper bound on the ROC curve of all positives $\bothpos$, whereas the set of contingency tables corresponding to the least upper bound on FPR form a lower bound on the ROC curve of $\bothpos$.

It is important to understand how these estimates correspond to bounds in ROC space. By computing $\theta$ as in Equation~\eqref{theta} to obtain the greatest lower bound on $\FPR(\latentpos,\ranksymb)$, the corresponding TPR is higher than $\TPR(\latentpos,\ranksymb)$. As such, the upper bound on the ROC curve is shifted upwards and to the left. Conversely, the lower bound on the ROC curve (based on the least upper bound on FPR at each rank, i.e. $\theta$ as in Equation~\eqref{theta-alt}) is shifted downward and to the right. This implies that the upper bound on the ROC curve completely dominates the curve of $\bothpos$ and the lower bound is completely dominated by the curve of $\bothpos$, provided that $\mathcal{T}_{lb}(\ranksymb) \leq \TPR(\latentpos,\ranksymb) \leq \mathcal{T}_{ub}(\ranksymb),\ \forall \ranksymb \in \{1,\ldots,|\overall|\}$.

{\bf Convergence properties} \label{convergence}
The convergence properties of our bounds are contingent on those of (a CI on) the empirical CDF:
via the strong law of large numbers the empirical CDF $\hat{F}_n(x)$ is a consistent pointwise estimator of the true CDF $F(x)$, converging uniformly for increasing $n$ \cite{van2000asymptotic}. 

Figure~\ref{fig:convergence} shows the convergence of the bounds on area under the curve for the estimated lower and upper bound of the ROC curve for increasing amounts of known positives in simulated rankings. The range of bounds depends on the width of the CI on rank CDF, which in turn depends on the number of known positives (higher is better) and the size of the total data set (lower is better).


{\bf PR curves} Given the contingency tables used to generate the least upper bound and greatest lower bound ROC curves, it is straightforward to construct the corresponding bounds in PR space. Each contingency table contains all the required information for generating a point in PR space. 

 A key result relating ROC and PR curves is that one curve dominates another in ROC space if and only if it also dominates in PR space~\citep{davis2006relationship}. Given this result, mapping the bounds we obtain for ROC curves to PR space directly yields (tight) bounds on the corresponding true PR curve. Since the upper bound in ROC space completely dominates the true curve, and the lower bound in ROC space is completely dominated by it, the same holds for the bounds on PR curves.


\newcommand{\covtype}{\texttt{covtype}\xspace}
\newcommand{\sensit}{\texttt{sensit}\xspace}

\newcommand{\resultcurves}[2]{
\begin{figure}[!h]
\centering
\subfigure[\covtype.]{\includegraphics[width=0.22\textwidth]{#1_covtype_pu_resvm.pdf}}\qquad
\subfigure[\sensit.]{\includegraphics[width=0.22\textwidth]{#1_sensit_2_semi_resvm.pdf}}\\
\caption{#2}
\label{fig:results-#1}
\end{figure}
}

\newcommand{\resultcurvesnew}[2]{
\begin{figure}[!h]
\centering
\subfigure[Rank CDF for \covtype.]{\includegraphics[width=0.2\textwidth]{#1_cdf.pdf}}\qquad
\subfigure[ROC curves for \covtype.]{\includegraphics[width=0.2\textwidth]{#1_roc.pdf}}\qquad
\subfigure[PR curves for \covtype.]{\includegraphics[width=0.2\textwidth]{#1_pr.pdf}}
\caption{#2}
\label{fig:results-#1}
\end{figure}
}

\section{Discussion and Recommendations}
Next, we discuss several issues related to using our approach in practice. 

\subsection{Determining $\hat{\pfrac}$ and its effect}

Our approach requires having an estimate $\hat{\pfrac}$ of $\pfrac$. There are many problems where $\pfrac$ is known from domain knowledge (e.g., calculated and published based on a data source you do not have access to), but explicit negatives are scarce or unavailable in the data under analysis. A real-world example where this is true is the task of predicting whether someone has diabetes from health insurance data \citep{claesen2015building}. In this context, some individuals are coded as having diabetes, but many diabetics are undiagnosed and hence it is wrong to assume that all unlabeled patients do not have diabetes. However, the incidence rate of diabetes is known and published in the medical literature. This type of situation characterizes many medical problems. If $\pfrac$ is not known from domain knowledge, then it could be estimated from data \citep{Elkan:2008:LCO:1401890.1401920, scottblanchard,scholkopf2001estimating}.



In either case, if $\hat{\pfrac}$ is not exact, the conditions of Lemma~\ref*{lemma-rank-fpr} are potentially violated where it is used within Theorem~\ref{main-theorem}. The effects of set size on FPR is characterized in Lemma~\ref*{lemma-size-fpr}, which will help us understand the effect of over or under estimating $\pfrac$.


\begin{replemma}{lemma-size-fpr}
Given two sets of positive labels $\pos_1$ and $\pos_2$ within an overall ranking $\overall$ and a rank $\ranksymb$, such that $\TPR(\pos_1,\ranksymb) = \TPR(\pos_2,\ranksymb)=t$ and $|\pos_1| > |\pos_2|$, then:
\begin{align*}
(a)\quad \FPR(\pos_2, r) < \tprsymb &\rightarrow \FPR(\pos_1, r) < \FPR(\pos_2, r), \\
(b)\quad \FPR(\pos_2, r) > \tprsymb &\rightarrow \FPR(\pos_1, r) > \FPR(\pos_2, r).
\end{align*}
(a) corresponds to a ranking and cutoff that is better than random (i.e. $\TPR(\pos,r) > \FPR(\pos,r)$).
\end{replemma}

\begin{figure}[!h]
  \centering
  \begin{minipage}[c]{0.6\textwidth}
	\centering
	\includegraphics[height=3.5cm]{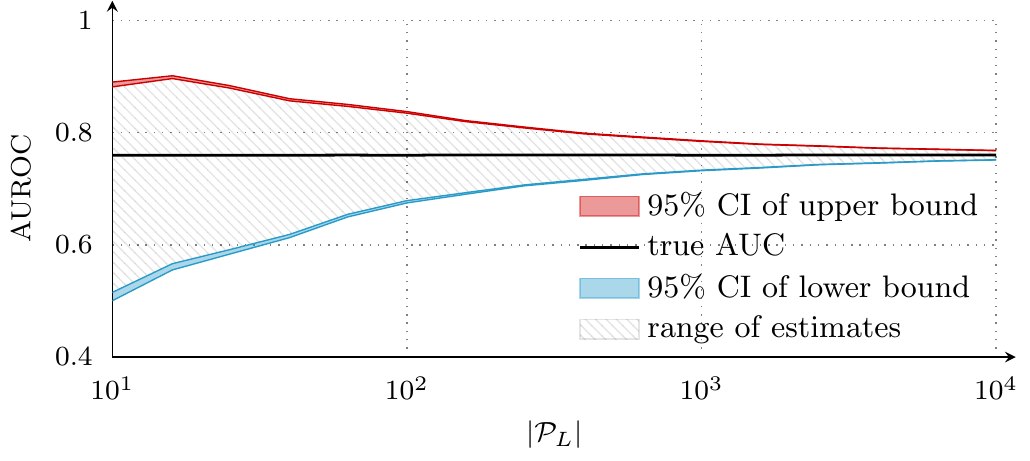}
	\vfill	
	\caption{The effect of $|\knownpos|$ on estimated AUC. Based on $|\unlabeled| = 100,000$, $\knownneg=\emptyset$ and $\hat{\pfrac}=\pfrac=0.2$. Bounds on rank CDF were obtained via bootstrap. The depicted confidence intervals are based on 200 repeated experiments.}
	\label{fig:convergence}
  \end{minipage}
  \hfill
  \begin{minipage}[c]{0.35\textwidth}
	\centering
	\includegraphics[height=3.5cm]{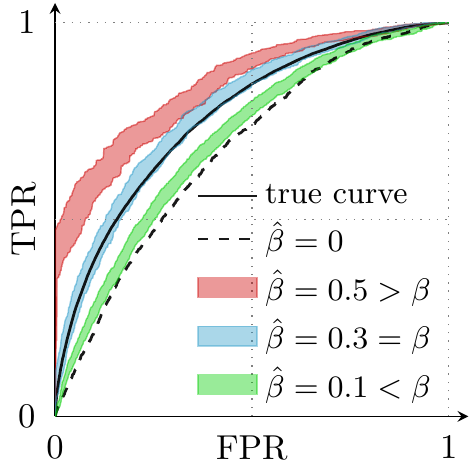}
	\vfill
	\caption{The effect of $\hat{\pfrac}$ on estimated ROC curves, based on 2,000 known positives, 100,000 unlabeled instances and $\pfrac=0.3$.}
	\label{fig:roc-ifv-beta}
  \end{minipage}
\end{figure}

Lemma~\ref*{lemma-size-fpr} has a large practical impact. If the ranking of $\knownpos$ is better than random, then over and under estimating $\hat{\pfrac}$ is useful to obtain a (loose) upper/lower bound on performance curves, respectively. In other words, given bounds or a CI on $\pfrac$, that is $\hat{\pfrac}_{lo} \leq \pfrac \leq \hat{\pfrac}_{up}$, we can use $\hat{\pfrac}_{lo}$ and $\hat{\pfrac}_{up}$ to estimate a lower and upper bound on the true ROC or PR curve. Bounds computed based on a CI for $\pfrac$ constitute a CI for the performance metric (at the same confidence level), assuming the rank CDF of $\latentpos$ is contained by the confidence band on the rank CDF. Tighter bounds on $\pfrac$ translate directly to tighter bounds on performance estimates. Finally, treating the full unlabeled set as negative underestimates the true performance, since $\hat{\pfrac}=0 < \pfrac$. The effect of varying $\hat{\pfrac}$ is shown in Figure~\ref{fig:roc-ifv-beta}.

\addtolength{\abovecaptionskip}{-5mm}
\addtolength{\abovecaptionskip}{5mm}


\subsection{Model selection}
Often evaluation metrics are used to select the best model from a set of candidates. If model A's ROC (PR) curve dominates model B's ROC (PR) curve, then for all $\pfrac$ model A is better than model B (leaving aside significance testing). However, in most cases one model does not dominate another model and there exists a point where the two curves cross. Surprisingly, the ordering in terms of both AUROC and AUPR are dependent on $\hat{\pfrac}$ when this happens. This means that the ordering of models according to these metrics can switch when $\hat{\pfrac}$ changes. Figure~\ref{rocauc-switch} depicts an example that illustrates this. This demonstrates that $\hat{\pfrac}$ can play a crucial role in model selection. In the likely event that the curves cross, it is important to look at the range of possible values for $\hat{\pfrac}$ that represent different operating conditions when selecting among different models.


A more formal explanation of why this occurs can be made based on partial derivatives of each entry of the partial contingency table and TPR, FPR and precision based on unlabeled instances to $\hat{\pfrac}$:\footnote{We made some simplifications, the details are described in Appendix~\ref*{partial}.}
\ifx
\begin{minipage}[t]{0.41\textwidth}
\begin{align*}
\frac{\partial \TP_U^r}{\partial \hat{\pfrac}} &= \mathcal{T}(r) \cdot |\unlabeled| &\geq 0, \\
\frac{\partial \FN_U^r}{\partial \hat{\pfrac}} &= \big(1 -\mathcal{T}(r)\big) \cdot |\unlabeled| &\geq 0, \\
\frac{\partial \FP_U^r}{\partial \hat{\pfrac}} &= -\mathcal{T}(r) \cdot |\unlabeled| &\leq 0, \\
\frac{\partial \TN_U^r}{\partial \hat{\pfrac}} &= \big(\mathcal{T}(r)-1\big) \cdot |\unlabeled| &\leq 0.
\end{align*}
\end{minipage}\hfill
\begin{minipage}[t]{0.53\textwidth}
\fi
\begin{equation}
\frac{\partial \TPR_U^r}{\partial \hat{\pfrac}} = 0,\quad 
\frac{\partial \FPR_U^r}{\partial \hat{\pfrac}} = \frac{|\topfun(\unlabeled, r)| - \mathcal{T}(r) \cdot |\unlabeled|}{(1-\hat{\pfrac})^2}, \quad
\frac{\partial \PREC_U^r}{\partial \hat{\pfrac}} = \frac{\mathcal{T}(r)\cdot|\unlabeled|}{|\topfun(\unlabeled,r)|} \geq 0. \label{eq:partials}
\end{equation}
The partial derivative of TPR is exactly 0 because our approach is based on rank CDFs (that is TPR at each rank). Interestingly, the partial derivatives of FPR and precision to $\hat{\pfrac}$ are dependent on the value of the rank CDF $\mathcal{T}(r)$ that is being used to infer surrogate positives. Since $\TPR$ is not a function of $\hat{\pfrac}$ and the partial derivatives of $\FPR$/precision to $\hat{\pfrac}$ are functions of $\mathcal{T}(r)$, distinct segments of an ROC/PR curve are moved differently when $\hat{\pfrac}$ changes, inducing a non-uniform scaling of AUC across the TPR range. Such scaling potentially changes the ordering of models based on AUC. 
\begin{figure}[!h]
\centering
  \begin{minipage}[c]{0.26\textwidth}
  \includegraphics[width=\textwidth]{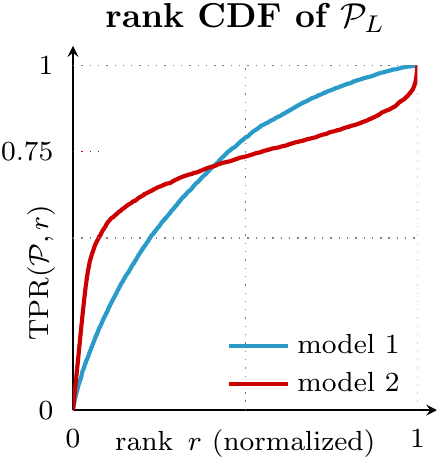}
  \end{minipage}\hfill
  \begin{minipage}[c]{0.26\textwidth}
  \includegraphics[width=\textwidth]{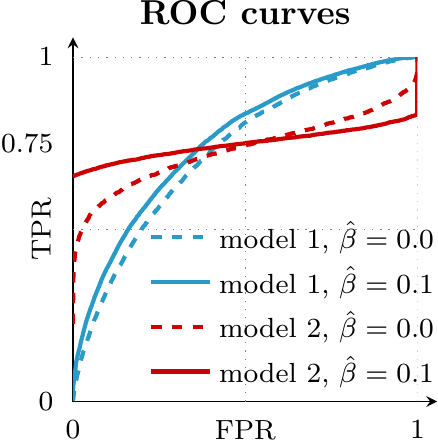}
  \end{minipage}
  \hfill
  \begin{minipage}[c]{0.43\textwidth}
\addtolength{\abovecaptionskip}{-7mm}
\caption[]{The effect of $\hat{\pfrac}$ on ROC curves. Setup: $|\unlabeled|=45,000$, $|\knownpos|=5,000$. 
\newline \ \newline 
Corresponding AUROC (best in bold):\\

{\centering
\begin{tabular}{ccc}
\toprule
estimated $\hat{\pfrac}$ 	& model 1 	& model 2 \\
\midrule
0.0		& $72.5\%$		& $\mathbf{73.2\%}$ \\
0.1		& $\mathbf{75.5\%}$	& $74.7\%$ \\
\bottomrule
\end{tabular}}
}\label{rocauc-switch}
\addtolength{\abovecaptionskip}{7mm}
  \end{minipage}
\end{figure}

\subsection{Empirical quality of the estimates} \label{empirical}
We illustrate the quality of our estimated bounds on ROC and PR curves using a model trained in a PU learning setting in \citep{claesen2014robust} on the \covtype data set \citep{Blackard00covtype}. The model was evaluated on a fully labeled test set of $20,000$ positive and $20,000$ negative examples. To estimate performance, we randomly selected $5\%$ of positive examples to serve as our labeled set and treated all other examples as unlabeled, which yields $|\knownpos|=1,000$, $|\unlabeled| = 39,000$ and $\pfrac\approx49\%$. We present ROC and PR curves with bounds for $\hat{\pfrac}=\pfrac$, $\hat{\pfrac}=0$, and a confidence interval $\hat{\pfrac}_{lo} = 0.8\pfrac \leq \hat{\pfrac} \leq \hat{\pfrac}_{up} = 1.2\pfrac$. Finally, as we have the ground truth, we present true curves as a reference.\footnote{Python code to reproduce all results (and modify the configuration) is available as supplementary material.}

Figure~\ref{fig:results-covtype} presents  the rank CDF and estimated bounds on ROC and PR curves. Figure~\ref{fig:results-covtype-cdf} shows the true rank CDF of $\latentpos$ along with an estimated $95\%$ CI on the rank CDF using the $\knownpos$ via a standard bootstrap approach with $2,000$ resamples. In this case, the CI contains the true rank CDF of latent positives.\footnote{The rank CDF of $\latentpos$ is unknown in practice, but assumed to be comparable to the rank CDF of $\knownpos$.} Figures~\ref{fig:results-covtype-roc} and~\ref{fig:results-covtype-pr} show that the bounds closely approximate the true performance curves. The estimated bounds are wider in PR space than in ROC space, particularly at low recall. Note that estimated PR curves are sensitive to the estimation error in $\hat{\pfrac}$, as precision is directly affected by class balance, limiting their usefulness if only a rough estimate of $\pfrac$ is available.



\begin{figure}[h]
\centering
\subfigure[Rank CDF.\label{fig:results-covtype-cdf}]{\includegraphics[width=0.25\textwidth]{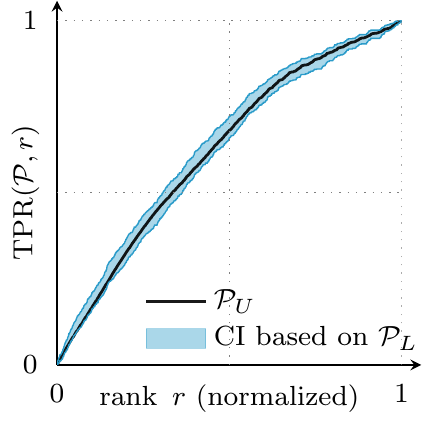}}\qquad
\subfigure[ROC curves.\label{fig:results-covtype-roc}]{\includegraphics[width=0.25\textwidth]{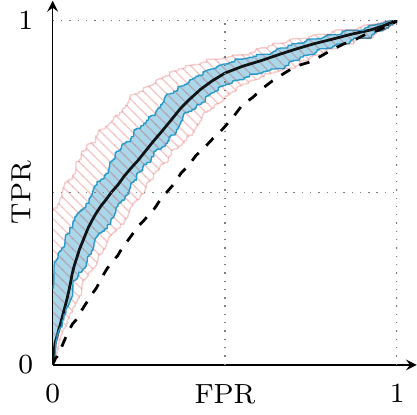}}\qquad
\subfigure[PR curves.\label{fig:results-covtype-pr}]{\includegraphics[width=0.25\textwidth]{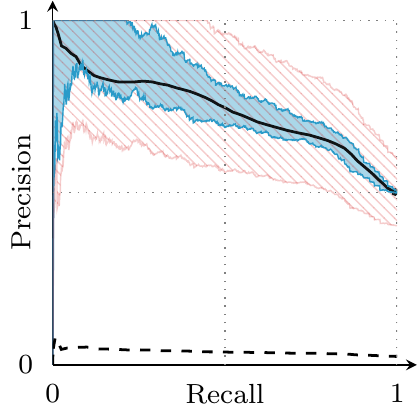}}
\addtolength{\abovecaptionskip}{-3mm}
\caption[]{Results for \covtype showing rank CDF, ROC and PR curves, with $\pfrac\approx 49\%$. \newline
Performance curve legend: 
\begin{tikzpicture}[baseline=-0.5ex] \draw [color=black, thick] (0, 0) -- (0.5, 0); \end{tikzpicture} true curve, 
\begin{tikzpicture}[baseline=-0.5ex] \draw [color=black, thick, dashed] (0, 0) -- (0.5, 0); \end{tikzpicture} $\hat{\pfrac}=0$, 
\begin{tikzpicture}[baseline=-0.5ex] \draw [color=cyan!80!black, opacity=0.8] (0, -0.2) rectangle (0.7, 0.3); 
\fill [color=cyan!80!black, opacity=0.4] (0, -0.2) rectangle (0.7, 0.3); \end{tikzpicture} $\hat{\pfrac}=\pfrac$ and 
\begin{tikzpicture}[baseline=-0.5ex] \draw [color=red!80!black, opacity=0.2, pattern=north west lines, pattern color=red!80!black] (0, -0.2) rectangle (0.7, 0.3); \end{tikzpicture} $0.8\pfrac \leq \hat{\pfrac} \leq 1.2\pfrac$.}
\label{fig:results-covtype}
\addtolength{\abovecaptionskip}{3mm}
\end{figure}

\subsection{Relative importance of known negatives compared to known positives}
As our approach can incorporate known negatives, a natural question is how their presence influences the estimates. In practice, a test set is of fixed size, so known negatives essentially reduce the size of the unlabeled subset, which in turn reduces the number of degrees of freedom in assigning surrogate positives. Using the same setup as in Subsection~\ref{empirical}, we varied the proportion of known positives and negatives and found known negatives provide some benefit, though this is small in practice. However, our approach can also be reversed given a large amount of negatives, that is flip known class labels, use $\bar{\pfrac}=1-\pfrac$ and adjust the resulting contingency tables accordingly, which can improve performance bounds. The benefits of known negatives are further discussed in Appendix~\ref*{knownneg}.





\section{Conclusion}
We presented an approach to construct contingency tables corresponding to a lower and upper bound on FPR using only partially labeled data, which enables computing many commonly used performance metrics in a semi-supervised setting. Our approach relies on knowing the fraction of latent positives in the unlabeled data, and we discussed its effect on determing the bounds and model selection. We have seen that our approach can yield good estimates in practice.  


\section*{Acknowledgments} 
\small{STADIUS members are supported by Flemish Government: FWO: projects:  G.0871.12N (Neural circuits), IWT: TBM-Logic Insulin(100793), TBM Rectal Cancer(100783), TBM IETA(130256); PhD grant \#111065, Industrial Research fund (IOF): IOF Fellowship 13-0260; iMinds Medical Information Technologies SBO 2015, ICON projects (MSIpad, MyHealthData)
VLK Stichting E. van der Schueren: rectal cancer; Federal Government: FOD: Cancer Plan 2012-2015 KPC-29-023 (prostate); COST: Action: BM1104: Mass Spectrometry Imaging. Jesse Davis is partially supported by the Research Fund KU Leuven (OT/11/051), EU FP7 Marie Curie Career Integration Grant (\#294068) and FWO-Vlaanderen (G.0356.12).}
\newpage

\bibliographystyle{unsrt}

\newpage
\section*{Supplementary material for ``\emph{Assessing Binary Classifiers Using Only Positive and Unlabeled Data}"}
\appendix
\section{Proofs} \label{proofs}

\begin{lemma} \label{lemma-rank-fpr}
Given a rank $\ranksymb$ and two disjoint subsets of positives $\pos_1$ and $\pos_2$ within an overall ranking $\overall$. If $|\pos_1|=|\pos_2|$ and $\TPR(\pos_1, \ranksymb) > \TPR(\pos_2,\ranksymb)$, then $\FPR(\pos_1,\ranksymb) < \FPR(\pos_2,\ranksymb)$ (see Figure~\ref{fig:lemma-rank-fpr}).

\begin{figure}[!h]
  \centering
  \includegraphics[width=0.6\textwidth]{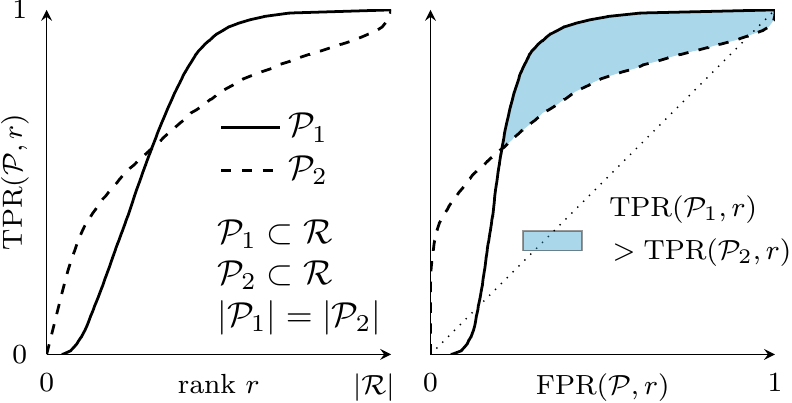}
  \caption{Illustration of Lemma~\ref*{lemma-rank-fpr}: higher TPR at a given rank $r$ implies lower FPR at $r$ for two positive sets of the same size.}
  \label{fig:lemma-rank-fpr}
\end{figure}

\textbf{Proof}: The numerator of FPR is the number of false positives, this is the number of positive predictions minus the number of true positives. Via Equations \eqref{fpr-def} and \eqref{tp-def}, this is $\ranksymb$ and $\TPR(\pos,\ranksymb)\cdot|\pos|$, respectively:
\begin{align}
\FPR(\pos, \ranksymb) &= \frac{\ranksymb - \TPR(\pos, r) \cdot |\pos|}{|\overall|-|\pos|} \label{tpr-fpr}.
\end{align}
Since $|\pos_1|=|\pos_2|$, the denominators of $\FPR(\pos_1,\ranksymb)$ and $\FPR(\pos_2,\ranksymb)$ are equal, so $\TPR(\pos_1,\ranksymb)>\TPR(\pos_2,\ranksymb) \leftrightarrow \FPR(\pos_1,\ranksymb) < \FPR(\pos_2,\ranksymb)$.\hfill$\blacksquare$
\end{lemma}

\begin{lemma} \label{lemma-rank-union}
Given a rank $\ranksymb$ and two disjoint sets of positives $\pos_1$ and $\pos_2$ in a ranking $\overall$ and $\bothpos=\pos_1\cup\pos_2$. If $\TPR(\pos_1,\ranksymb) = \tprsymb_1 < \TPR(\pos_2, \ranksymb) = \tprsymb_2$ then $\TPR(\pos_1,\ranksymb) < \TPR(\bothpos,\ranksymb) < \TPR(\pos_2, \ranksymb)$ (see Figure~\ref{fig:lemma-rank-union}).

\begin{figure}[!h]
  \centering
  \includegraphics[width=0.6\textwidth]{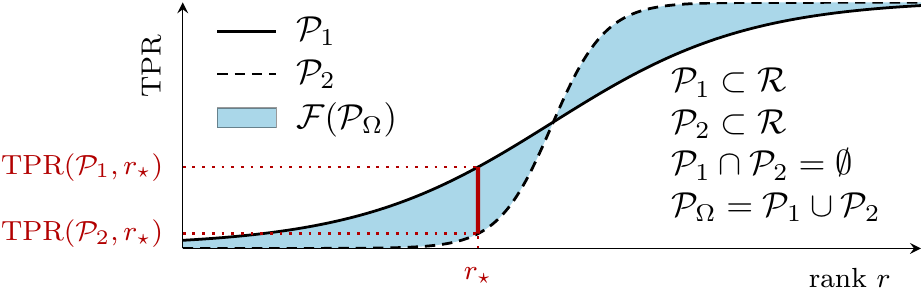}
  \caption{Illustration of Lemma~\ref*{lemma-rank-union}: $\mathcal{F}(\cdot)$ denotes feasible region. The rank distribution of the union $\bothpos$ of two sets of positives $\pos_1$ and $\pos_2$ lies between their respective rank distributions.} 
  \label{fig:lemma-rank-union}
\end{figure}

\textbf{Proof}: write $\TPR(\bothpos,\ranksymb)$ in terms of $\tprsymb_1$ and $\tprsymb_2$:
\begin{equation}
\TPR(\bothpos,\ranksymb) = \frac{\tprsymb_1 \cdot|\pos_1| + \tprsymb_2\cdot|\pos_2|}{|\pos_1|+|\pos_2|}. \label{tpr-union}
\end{equation}
since $\tprsymb_1 < \tprsymb_2$, we get $\tprsymb_1 < \TPR(\bothpos,\ranksymb) < \tprsymb_2$.\hfill$\blacksquare$
\end{lemma}

\begin{lemma} \label{lemma-glb}
Minimizing $\TPR(\surrpos, \ranksymb)$ in Equation~\eqref{eq:surrpos-def} of Theorem~\ref{main-theorem} ensures $\FPR(\bothposapprox, \ranksymb)$ is the greatest achievable lower bound on $\FPR(\bothpos, \ranksymb)$ given $\pfrac$, $\mathcal{T}_{ub}(\ranksymb)$, $\overall$ and $\unlabeled$. 

\textbf{Proof} (by contradiction): suppose another set of surrogate positives $\surrposbullet \subset \unlabeled$ exists with $|\surrposbullet|=\pfrac\cdot|\unlabeled|$, such that $\surrposbullet\neq\surrpos$, and $\TPR(\surrposbullet,\ranksymb) \geq \mathcal{T}_{ub}(\ranksymb)$ and for $\bothposbullet = \knownpos\cup\surrposbullet$: 
\begin{equation*}
\FPR(\bothposapprox,\ranksymb) < \FPR(\bothposbullet,\ranksymb) \leq \FPR(\bothpos,\ranksymb).
\end{equation*}
Via Corollary~\ref{corollary-rank-union} this implies $\TPR(\surrposbullet,\ranksymb) < \TPR(\surrpos,\ranksymb)$, which contradicts the definition of $\surrpos$ (Eq.~\eqref{eq:surrpos-def}).\hfill$\blacksquare$
\end{lemma}

\begin{corollary} \label{corollary-rank-union}
Given a rank $\ranksymb$ and three sets of positives $\pos_A$, $\pos_B$ and $\pos_C$ within a ranking $\overall$ such that $\pos_A\cap\pos_B=\emptyset$ and $\pos_A\cap\pos_C=\emptyset$ and $|\pos_B|=|\pos_C|$, then
\begin{equation*}
\TPR(\pos_B,\ranksymb) = \tprsymb_B < \TPR(\pos_C,\ranksymb) = \tprsymb_C \quad\leftrightarrow\quad \TPR(\pos_A\cup \pos_B,\ranksymb) < \TPR(\pos_A\cup\pos_C,\ranksymb).
\end{equation*}
\textbf{Proof}: all terms are equal for $\TPR(\pos_A\cup \pos_B,\ranksymb)$ and $\TPR(\pos_A\cup \pos_C,\ranksymb)$ except $\tprsymb_B < \tprsymb_C$ in Eq.~\eqref{tpr-union}.
\end{corollary}

\begin{lemma} \label{lemma-size-fpr}
Given two sets of positive labels $\pos_1$ and $\pos_2$ within an overall ranking $\overall$ and a rank $\ranksymb$, such that $\TPR(\pos_1,\ranksymb) = \TPR(\pos_2,\ranksymb)=t$ and $|\pos_1| > |\pos_2|$, then:
\begin{align*}
\begin{tikzpicture}[baseline=-0.5ex] \node [draw=none, fill=cyan!80!black, fill opacity=0.4, text opacity=0.9, rounded corners] (blah) {$(a)$}; \end{tikzpicture}
\quad \FPR(\pos_2, r) < \tprsymb &\rightarrow \FPR(\pos_1, r) < \FPR(\pos_2, r), \\
\begin{tikzpicture}[baseline=-0.5ex] \node [draw=none, fill=red!80!black, fill opacity=0.4, text opacity=0.9, rounded corners] (blah) {$(b)$}; \end{tikzpicture}
\quad \FPR(\pos_2, r) > \tprsymb &\rightarrow \FPR(\pos_1, r) > \FPR(\pos_2, r).
\end{align*}
(a) corresponds to a ranking and cutoff that is better than random (i.e. $\TPR(\pos,r) > \FPR(\pos,r)$) whereas (b) corresponds to a ranking and cutoff that is worse than random.

\begin{figure}[!h]
  \centering
  \includegraphics[width=0.6\textwidth]{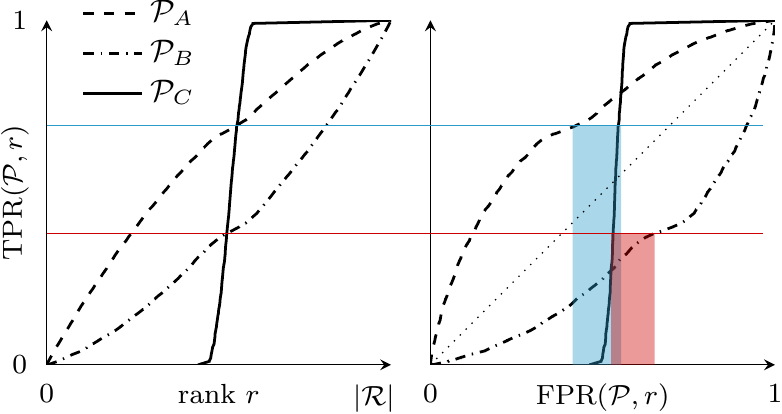}
  \caption[]{Illustration of Lemma~\ref*{lemma-size-fpr}, with $\pos_A \subset \overall$, $\pos_B \subset \overall$, $\pos_C \subset \overall$, \begin{tikzpicture}[baseline=-0.5ex] \node [draw=none, fill=cyan!80!black, fill opacity=0.4, text opacity=0.9, rounded corners] (blah) {$|\pos_A| > |\pos_C|$}; \end{tikzpicture} and \begin{tikzpicture}[baseline=-0.5ex] \node [draw=none, fill=red!80!black, fill opacity=0.4, text opacity=0.9, rounded corners] (blah) {$|\pos_B| > |\pos_C|$}; \end{tikzpicture}. If two sets of positives $\pos_1$ and $\pos_2$ achieve a given TPR at the same rank $r$, e.g. $\TPR(\pos_1,r)=\TPR(\pos_2,r)$ and $|\pos_1| > |\pos_2|$ then $\FPR(\pos_1,r) < \FPR(\pos_2,r)$ if $\FPR(\pos_2,r) < \TPR(\pos_2,r)$ and otherwise $\FPR(\pos_1,r) > \FPR(\pos_2,r)$.} 
  \label{fig:lemma-size-fpr}
\end{figure}

\textbf{Proof}: take the derivative of FPR to $|\pos|$ while fixing $\ranksymb$, based on Equation~\eqref{tpr-fpr}:
\begin{align}
\frac{d\FPR(\pos, \ranksymb)}{d|\pos|} &= \frac{\ranksymb-\tprsymb\cdot|\overall|}{(|\overall|-|\pos|)^2}, \nonumber \\
&= \frac{\ranksymb-\tprsymb\cdot|\pos| - \tprsymb \cdot|\overall-\pos|}{(|\overall|-|\pos|)^2}. \label{FPR_ifv_gamma}
\end{align}
$r-\tprsymb\cdot|\pos|$ is the number of negatives in the top ranking (false positives) and $\tprsymb\cdot|\overall - \pos|$ is the number of false positives at $\FPR=\tprsymb$. The derivative is negative if the $\FPR$ is below $\tprsymb$ and vice versa, therefore if the ranking is better than random ($\TPR = \tprsymb > \FPR$), increasing $|\pos|$ leads to a lower $\FPR$ at rank $\ranksymb$ and vice versa. \hfill $\blacksquare$
\end{lemma}

\section{Effect of $\hat{\pfrac}$ on contingency table entries and common performance metrics} \label{partial}
To study the effect of imprecise estimates of $\pfrac$, we start by computing partial derivatives of each entry of the partial contingency table based on unlabeled instances to $\hat{\pfrac}$ (see Section~\ref{contingency}). Subsequently, we will compute partial derivatives of TPR, FPR and precision to $\hat{\pfrac}$ to describe the effect of estimating $\pfrac$ on (area under) ROC and PR curves.

For ease of notation, we base all subsequent calculations on $\tilde{\theta} = \hat{\pfrac} \mathcal{T}(r) \cdot |\unlabeled| \approx \theta$ which ignores the discrete effect of rounding in the real definition of $\theta$ (Eq.~\ref{theta}). We additionally assume it is possible to assign the desired amount $\tilde{\theta}$ of surrogate positives in $\topfun(\unlabeled, r)$, which holds for ranks $r$ that are not too close to the top or bottom of $\overall$, given reasonable values of $\hat{\pfrac}$ and CDF bounds $\mathcal{T}(r)$.\footnote{$\mathcal{T}(r)$ represents a bound on rank CDF, that is either $\mathcal{T}_{lb}(r)$ or $\mathcal{T}_{ub}(r)$ as used in the manuscript.} If this does not hold, that is when there is clipping in Eq.~\ref{eq:surrpos-conditions}, then (small) changes in $\hat{\pfrac}$ do not affect $\TP_U^r$ and hence the partial derivatives of \emph{all} entries in the contingency table to $\hat{\pfrac}$ are effectively 0.

Given these simplifications, the partial contingency table based on unlabeled instances becomes:
\begin{align*}
\TP_U^r &= \tilde{\theta} = \hat{\pfrac}\mathcal{T}(r) \cdot |\unlabeled| \\
\FN_U^r &= |\surrpos| - \TP_U^r = \hat{\pfrac}\cdot |\unlabeled| - \hat{\pfrac}\mathcal{T}(r) \cdot |\unlabeled| = \hat{\pfrac} \big(1-\mathcal{T}(r)\big) \cdot|\unlabeled| \nonumber \\
\FP_U^r &= |\topfun(\unlabeled, r)| - \TP_U^r = |\topfun(\unlabeled, r)| - \hat{\pfrac}\mathcal{T}(r) \cdot |\unlabeled|, \nonumber \\
\TN_U^r &= |\unlabeled| - |\surrpos| - \FP_U^r = |\unlabeled| - \hat{\pfrac}\cdot|\unlabeled| - |\topfun(\unlabeled, r)| + \hat{\pfrac}\mathcal{T}(r) \cdot |\unlabeled|, \\
	&= \big(1-\hat{\pfrac} + \hat{\pfrac}\mathcal{T}(r)\big) \cdot |\unlabeled| - |\topfun(\unlabeled, r)|.
\end{align*}
The partial derivatives of each entry of the partial contingency table then become:
\begin{align*}
\frac{\partial \TP_U^r}{\partial \hat{\pfrac}} &= \mathcal{T}(r) \cdot |\unlabeled| \geq 0, & &\frac{\partial \FP_U^r}{\partial \hat{\pfrac}} = -\mathcal{T}(r) \cdot |\unlabeled| \leq 0, \\
\frac{\partial \FN_U^r}{\partial \hat{\pfrac}} &= \big(1 -\mathcal{T}(r)\big) \cdot |\unlabeled| \geq 0, & &\frac{\partial \TN_U^r}{\partial \hat{\pfrac}} = \big(\mathcal{T}(r)-1\big) \cdot |\unlabeled| \leq 0.
\end{align*}

Partial derivatives for TPR, TPR and precision are a little more involved:
\begin{align}
\frac{\partial \TPR_U^r}{\partial \hat{\pfrac}} &= \frac{\frac{\partial \TP_U^r}{\partial \hat{\pfrac}} |\surrpos| - \TP_U^r \frac{\partial |\surrpos|}{\partial \hat{\pfrac}}}{|\surrpos|^2} 
= \frac{\mathcal{T}(r)\hat{\pfrac}\cdot|\unlabeled|^2-\mathcal{T}(r)\hat{\pfrac} |\unlabeled|^2}{\hat{\pfrac}^2|\unlabeled|^2} 
= \frac{\mathcal{T}(r)-\mathcal{T}(r)}{\hat{\pfrac}} = 0 \label{eq:dtpr} \\
\frac{\partial \FPR_U^r}{\partial \hat{\pfrac}} &= \frac{\frac{\partial \FP_U^r}{\partial \hat{\pfrac}} \cdot (|\unlabeled|-|\surrpos|) - \FP_U^r \frac{\partial (|\unlabeled|-|\surrpos|)}{\partial \hat{\pfrac}}}{(|\unlabeled|-|\surrpos|)^2} = \frac{-\mathcal{T}(r) \cdot |\unlabeled| \cdot (|\unlabeled|-|\surrpos|)+\FP_U^r \cdot |\unlabeled|}{(|\unlabeled|-|\surrpos|)^2} \nonumber \\
&=\frac{-\mathcal{T}(r) (1-\hat{\pfrac}) \cdot |\unlabeled|^2 + \FP_U^r \cdot |\unlabeled|}{(1-\hat{\pfrac})^2 \cdot |\unlabeled|^2} 
=\frac{-\mathcal{T}(r)}{1-\hat{\pfrac}} + \frac{(|\topfun(\unlabeled, r)|-\hat{\pfrac}\mathcal{T}(r) \cdot |\unlabeled|) \cdot |\unlabeled|}{(1-\hat{\pfrac})^2 \cdot |\unlabeled|^2}  \nonumber \\
&=\frac{-\mathcal{T}(r)}{(1-\hat{\pfrac})^2} + \frac{|\topfun(\unlabeled, r)|}{(1-\hat{\pfrac})^2\cdot|\unlabeled|} 
= \frac{|\topfun(\unlabeled, r)| - \mathcal{T}(r) \cdot |\unlabeled|}{(1-\hat{\pfrac})^2} \label{eq:dfpr} \\
\frac{\partial \PREC_U^r}{\partial \hat{\pfrac}} &= \frac{\frac{\partial \TP_U^r}{\partial \hat{\pfrac}} \cdot (\TP_U^r+\FP_U^r) - \TP_U^r \frac{\partial (\TP_U^r+\FP_U^r)}{\partial \hat{\pfrac}}}{(\TP_U^r+\FP_U^r)^2} \nonumber \\ 
&= \frac{\mathcal{T}(r)\cdot|\unlabeled|\cdot(\TP_U^r + \FP_U^r)}{(\TP_U^r + \FP_U^r)^2} 
= \frac{\mathcal{T}(r)\cdot|\unlabeled|}{(\TP_U^r + \FP_U^r)} = \frac{\mathcal{T}(r)\cdot|\unlabeled|}{|\topfun(\unlabeled,r)|} \geq 0 \label{eq:dprec} 
\end{align}

Both $\partial \FPR_U^r / \partial \hat{\pfrac}$ and $\partial \PREC_U^r / \partial \hat{\pfrac}$ are a function of $\mathcal{T}(r)$, while $\partial \FPR_U^r / \partial \hat{\pfrac} = 0$. This implies that the ordering of rankings in terms of area under the ROC curve can change when the estimate of $\pfrac$ changes, as proven by example in Figure~\ref{rocauc-switch}.

\section{The effect of the fraction of known positives, known negatives and $\hat{\pfrac}$} \label{knownneg}
Known negatives can be incorporated in our approach as described in Section~\ref{contingency}. Given a fixed ranking $\overall$, having known negatives essentially reduces the size of the unlabeled subset $\unlabeled$, which in turn reduces the number of degrees of freedom in assigning surrogate positives. As such, known negatives provide some benefit, though this is small in practice. Table~\ref{parameffects-posonly} illustrates the effect of increasing amounts of known positives and known negatives: known positives significantly tighten bounds on AUROC, while known negatives only do so marginally (cfr. bounds with $10\%$ known positives and $40/60/80\%$ known negatives).

\def\realauc{0.765}

\newcommand{\plotauc}[2]{
\begin{tikzpicture}[baseline=1ex]
    \begin{axis}[hide axis,clip=false,
	xmin=0.67,xmax=0.87,xlabel={X},
	ymin=-1,ymax=1,
	x=12cm, y=0.8em,
	]

\addplot[black, very thick, dashed] coordinates {(\realauc, -1) (\realauc, 1)};
\addplot[gray, thick, dashed] coordinates {(0.67, -1) (0.67, 1)};
\addplot[gray, thick, dashed] coordinates {(0.87, -1) (0.87, 1)};
\addplot[black, dotted] coordinates {(0.73, -1) (0.73, 1)};
\addplot[black, dotted] coordinates {(0.75, -1) (0.75, 1)};
\addplot[black, dotted] coordinates {(0.77, -1) (0.77, 1)};
\addplot[black, dotted] coordinates {(0.79, -1) (0.79, 1)};
\addplot[black, dotted] coordinates {(0.81, -1) (0.81, 1)};
\addplot[gray, dotted] coordinates {(0.71, -1) (0.71, 1)};
\addplot[gray, dotted] coordinates {(0.69, -1) (0.69, 1)};
\addplot[gray, dotted] coordinates {(0.83, -1) (0.83, 1)};
\addplot[gray, dotted] coordinates {(0.85, -1) (0.85, 1)};
\draw [color=cyan!80!black, opacity=0.8] (axis cs: #1, -0.7) rectangle (axis cs: #2, 0.6);
\fill [color=cyan!80!black, opacity=0.4, fill] (axis cs: #1, -0.7) rectangle (axis cs: #2, 0.6);
    \end{axis}
\end{tikzpicture}
}
\newcommand{\plotauccolor}[3]{
\begin{tikzpicture}[baseline=1ex]
    \begin{axis}[hide axis,clip=false,
	xmin=0.67,xmax=0.87,xlabel={X},
	ymin=-1,ymax=1,
	x=12cm, y=0.8em,
	]

\addplot[black, very thick, dashed] coordinates {(\realauc, -1) (\realauc, 1)};
\addplot[gray, thick, dashed] coordinates {(0.67, -1) (0.67, 1)};
\addplot[gray, thick, dashed] coordinates {(0.87, -1) (0.87, 1)};
\addplot[black, dotted] coordinates {(0.73, -1) (0.73, 1)};
\addplot[black, dotted] coordinates {(0.75, -1) (0.75, 1)};
\addplot[black, dotted] coordinates {(0.77, -1) (0.77, 1)};
\addplot[black, dotted] coordinates {(0.79, -1) (0.79, 1)};
\addplot[black, dotted] coordinates {(0.81, -1) (0.81, 1)};
\addplot[gray, dotted] coordinates {(0.71, -1) (0.71, 1)};
\addplot[gray, dotted] coordinates {(0.69, -1) (0.69, 1)};
\addplot[gray, dotted] coordinates {(0.83, -1) (0.83, 1)};
\addplot[gray, dotted] coordinates {(0.85, -1) (0.85, 1)};
\draw [color=#3!80!black, opacity=0.8] (axis cs: #1, -0.7) rectangle (axis cs: #2, 0.6);
\fill [color=#3!80!black, opacity=0.4, fill] (axis cs: #1, -0.7) rectangle (axis cs: #2, 0.6);
    \end{axis}
\end{tikzpicture}
}
\newcommand{\plotauclabels}{
\begin{tikzpicture}[baseline=0.8ex]
    \begin{axis}[hide axis,clip=false,
	xmin=0.67,xmax=0.87,xlabel={X},
	ymin=-1,ymax=1,
	x=12cm, y=0.8em,
	]
\node [gray] at (axis cs: 0.67, 0) {$67\%$};
\node [black] at (axis cs: 0.763, 0) {$76.8\%$};
\node [gray] at (axis cs: 0.87, 0) {$87\%$};
\addplot[gray, dotted] coordinates {(0.7068, -1) (0.7068, 1)};
\addplot[gray, dotted] coordinates {(0.8268, -1) (0.8268, 1)};
\addplot[black, dotted] coordinates {(0.8068, -1) (0.8068, 1)};
    \end{axis}
\end{tikzpicture}
}

However, when the number of known negatives is large, it may be useful to reverse our approach, i.e., start from the rank distribution of known negatives. To do so, we can essentially flip all known class labels, use $\bar{\pfrac}=1-\pfrac$ and adjust the resulting contingency tables accordingly.

Table~\ref{parameffects-both} shows bounds when based on known positives or known negatives (whichever are tightest). It is important to see that $|\knownneg| > |\knownpos|$ does not guarantee that performance bounds based on known negatives are tighter, because $\pfrac$ also affects the bounds. When computing performance bounds based on known negatives, overestimating $\hat{\pfrac}$ leads to underestimated bounds (since we use $\bar{\pfrac}=1-\hat{\pfrac}$) and vice versa. The effect of errors in $\hat{\pfrac}$ is opposite in bounds based on $\knownneg$.

Hence, bounds on performance metrics can be computed based primarily on known positives $\knownpos$ \emph{or} known negatives $\knownneg$. The width of the bounds depends on the combination of $|\knownpos|$ (or $|\knownneg|$) and $\pfrac$ (or $\bar{\pfrac}$) in a nontrivial way: depending on $\pfrac$, it is possible to obtain wider bounds based on known negatives, even if $|\knownneg| > |\knownpos|$ (or vice versa). In practice, we can estimate metrics based on $\knownpos$ and $\knownneg$ separately and then use whichever yields the tightest bounds, as shown in Table~\ref{parameffects-both}.

\begin{table}[!h]
\begin{tabular}{ccccc|c|c}
\toprule
\multicolumn{3}{c}{configuration} & & \multicolumn{3}{c}{bounds on area under the ROC curve (true AUROC=$76.8\%$)} \\ \cline{1-3} \cline{5-7}
$\frac{|\knownpos|}{|\bothpos|}$	& $\frac{|\knownneg|}{|\mathcal{N}_\Omega|}$ & $\pfrac$ & & $\hat{\pfrac}\ /\ \pfrac = 0.8$ & $\hat{\pfrac}\ /\ \pfrac = 1.0$  & $\hat{\pfrac}\ /\ \pfrac = 1.2$ \\ 
\midrule
10				&	0	& 15	& & \plotauc{0.7099}{0.8063}  & \plotauc{0.7163}{0.8182}  & \plotauc{0.7235}{0.8303} \\
				&	20	& 18	& & \plotauc{0.7099}{0.8062}  & \plotauc{0.7163}{0.8180}  & \plotauc{0.7235}{0.8299} \\
				&	40	& 23	& & \plotauc{0.7099}{0.8060}  & \plotauc{0.7163}{0.8176}  & \plotauc{0.7235}{0.8289} \\
				&	60	& 31	& & \plotauc{0.7099}{0.8059}  & \plotauc{0.7163}{0.8059}  & \plotauc{0.7235}{0.8270} \\
				&	80	& 47	& & \plotauc{0.7099}{0.8055}  & \plotauc{0.7163}{0.8055}  & \plotauc{0.7235}{0.8160} \\
 & & & & \plotauclabels & \plotauclabels  & \plotauclabels \\
30				&	0	& 12	& & \plotauc{0.7328}{0.7732}  & \plotauc{0.7378}{0.7823}  & \plotauc{0.7434}{0.7915} \\
				&	20	& 15	& & \plotauc{0.7328}{0.7732}  & \plotauc{0.7378}{0.7823}  & \plotauc{0.7434}{0.7915} \\
				&	40	& 19	& & \plotauc{0.7328}{0.7732}  & \plotauc{0.7378}{0.7823}  & \plotauc{0.7434}{0.7914} \\
				&	60	& 26	& & \plotauc{0.7328}{0.7732}  & \plotauc{0.7378}{0.7823}  & \plotauc{0.7434}{0.7914} \\
				&	80	& 41	& & \plotauc{0.7328}{0.7732}  & \plotauc{0.7378}{0.7822}  & \plotauc{0.7434}{0.7907} \\
 & & & & \plotauclabels & \plotauclabels  & \plotauclabels \\
50				&	0	& 9	& & \plotauc{0.7455}{0.7682}  & \plotauc{0.7502}{0.7739}  & \plotauc{0.7573}{0.7807} \\
				&	20	& 11	& & \plotauc{0.7455}{0.7682}  & \plotauc{0.7542}{0.7779}  & \plotauc{0.7573}{0.7807} \\
				&	40	& 14	& & \plotauc{0.7455}{0.7682}  & \plotauc{0.7542}{0.7779}  & \plotauc{0.7573}{0.7807} \\
				&	60	& 20	& & \plotauc{0.7455}{0.7682}  & \plotauc{0.7542}{0.7779}  & \plotauc{0.7573}{0.7807} \\
				&	80	& 33	& & \plotauc{0.7455}{0.7682}  & \plotauc{0.7542}{0.7779}  & \plotauc{0.7573}{0.7804} \\
 & & & & \plotauclabels & \plotauclabels  & \plotauclabels \\
70				&	0	& 6	& & \plotauc{0.7533}{0.7646}  & \plotauc{0.7594}{0.7727}  & \plotauc{0.7676}{0.7778} \\
				&	20	& 7	& & \plotauc{0.7533}{0.7646}  & \plotauc{0.7594}{0.7727}  & \plotauc{0.7676}{0.7778} \\
				&	40	& 9	& & \plotauc{0.7533}{0.7646}  & \plotauc{0.7594}{0.7727}  & \plotauc{0.7676}{0.7778} \\
				&	60	& 13	& & \plotauc{0.7533}{0.7646}  & \plotauc{0.7594}{0.7727}  & \plotauc{0.7676}{0.7778} \\
				&	80	& 23	& & \plotauc{0.7533}{0.7646}  & \plotauc{0.7594}{0.7727}  & \plotauc{0.7676}{0.7778} \\
\bottomrule
\end{tabular}
\caption[]{Estimated bounds on AUROC under different configurations. The total data set comprises $2,000$ positives and $10,000$ negatives. We varied the fraction of known positives and known negatives, which also implies changing $\pfrac$. All entries in the table are in percentages. We used three estimates for $\hat{\pfrac}$, namely an underestimate, the correct value and an overestimate (left to right). \newline
Legend:
\begin{tikzpicture}[baseline=-0.5ex] \draw [color=black, thick, dashed] (0, 0) -- (0.5, 0); \end{tikzpicture} true AUROC, 
\begin{tikzpicture}[baseline=-0.5ex] \draw [color=cyan!80!black, opacity=0.8] (0, -0.15) rectangle (0.7, 0.25); 
\fill [color=cyan!80!black, opacity=0.4] (0, -0.15) rectangle (0.7, 0.25); \end{tikzpicture} bounds based on known positives.
}
\label{parameffects-posonly}
\end{table}

\begin{table}[!h]
\begin{tabular}{ccccc|c|c}
\toprule
\multicolumn{3}{c}{configuration} & & \multicolumn{3}{c}{bounds on area under the ROC curve (true AUROC=$76.8\%$)} \\ \cline{1-3} \cline{5-7}
$\frac{|\knownpos|}{|\bothpos|}$	& $\frac{|\knownneg|}{|\mathcal{N}_\Omega|}$ & $\pfrac$ & & $\hat{\pfrac}\ /\ \pfrac = 0.8$ & $\hat{\pfrac}\ /\ \pfrac = 1.0$  & $\hat{\pfrac}\ /\ \pfrac = 1.2$ \\ 
\midrule
10 & 0 & 15 & & \plotauccolor{0.7081}{0.8044}{cyan} & \plotauccolor{0.7143}{0.8160}{cyan} & \plotauccolor{0.7214}{0.8277}{cyan} \\
 & 20 & 18 & & \plotauccolor{0.7081}{0.8043}{cyan} & \plotauccolor{0.7143}{0.8159}{cyan} & \plotauccolor{0.7214}{0.8274}{cyan} \\
 & 40 & 23 & & \plotauccolor{0.7950}{0.8724}{red} & \plotauccolor{0.7420}{0.8215}{red} & \plotauccolor{0.7067}{0.7742}{red} \\
 & 60 & 31 & & \plotauccolor{0.8106}{0.8531}{red} & \plotauccolor{0.7564}{0.8003}{red} & \plotauccolor{0.7182}{0.7538}{red} \\
 & 80 & 47 & & \plotauccolor{0.8072}{0.8287}{red} & \plotauccolor{0.7534}{0.7729}{red} & \plotauccolor{0.7165}{0.7301}{red} \\
 & & & & \plotauclabels & \plotauclabels  & \plotauclabels \\
30 & 0 & 12 & & \plotauccolor{0.7367}{0.7803}{cyan} & \plotauccolor{0.7417}{0.7896}{cyan} & \plotauccolor{0.7474}{0.7989}{cyan} \\
 & 20 & 14 & & \plotauccolor{0.7367}{0.7802}{cyan} & \plotauccolor{0.7417}{0.7896}{cyan} & \plotauccolor{0.7474}{0.7988}{cyan} \\
 & 40 & 18 & & \plotauccolor{0.7367}{0.7802}{cyan} & \plotauccolor{0.7417}{0.7895}{cyan} & \plotauccolor{0.7474}{0.7987}{cyan} \\
 & 60 & 25 & & \plotauccolor{0.7964}{0.8347}{red} & \plotauccolor{0.7567}{0.7996}{red} & \plotauccolor{0.7256}{0.7638}{red} \\
 & 80 & 41 & & \plotauccolor{0.7935}{0.8125}{red} & \plotauccolor{0.7539}{0.7727}{red} & \plotauccolor{0.7235}{0.7386}{red} \\
 & & & & \plotauclabels & \plotauclabels  & \plotauclabels \\
50 & 0 & 9 & & \plotauccolor{0.7488}{0.7706}{cyan} & \plotauccolor{0.7524}{0.7773}{cyan} & \plotauccolor{0.7564}{0.7841}{cyan} \\
 & 20 & 11 & & \plotauccolor{0.7488}{0.7706}{cyan} & \plotauccolor{0.7524}{0.7773}{cyan} & \plotauccolor{0.7564}{0.7841}{cyan} \\
 & 40 & 14 & & \plotauccolor{0.7488}{0.7706}{cyan} & \plotauccolor{0.7524}{0.7773}{cyan} & \plotauccolor{0.7564}{0.7840}{cyan} \\
 & 60 & 20 & & \plotauccolor{0.7488}{0.7706}{cyan} & \plotauccolor{0.7524}{0.7773}{cyan} & \plotauccolor{0.7564}{0.7840}{cyan} \\
 & 80 & 33 & & \plotauccolor{0.7807}{0.7992}{red} & \plotauccolor{0.7537}{0.7724}{red} & \plotauccolor{0.7313}{0.7475}{red} \\
 & & & & \plotauclabels & \plotauclabels  & \plotauclabels \\
70 & 0 & 5 & & \plotauccolor{0.7554}{0.7667}{cyan} & \plotauccolor{0.7576}{0.7709}{cyan} & \plotauccolor{0.7599}{0.7751}{cyan} \\
 & 20 & 6 & & \plotauccolor{0.7554}{0.7667}{cyan} & \plotauccolor{0.7576}{0.7709}{cyan} & \plotauccolor{0.7599}{0.7751}{cyan} \\
 & 40 & 9 & & \plotauccolor{0.7554}{0.7667}{cyan} & \plotauccolor{0.7576}{0.7709}{cyan} & \plotauccolor{0.7599}{0.7751}{cyan} \\
 & 60 & 13 & & \plotauccolor{0.7554}{0.7667}{cyan} & \plotauccolor{0.7576}{0.7709}{cyan} & \plotauccolor{0.7599}{0.7751}{cyan} \\
 & 80 & 23 & & \plotauccolor{0.7554}{0.7667}{cyan} & \plotauccolor{0.7576}{0.7709}{cyan} & \plotauccolor{0.7599}{0.7751}{cyan} \\
\bottomrule
\end{tabular}
\caption[]{Estimated bounds on AUROC under different configurations. The total data set comprises $2,000$ positives and $10,000$ negatives. We varied the fraction of known positives and known negatives, which also implies changing $\pfrac$. All entries in the table are in percentages. We used three estimates for $\hat{\pfrac}$, namely an underestimate, the correct value and an overestimate (left to right). In this table, we computed bounds based on known positives and known negatives (separately) and report the tightest confidence interval each time. \newline
Legend: 
\begin{tikzpicture}[baseline=-0.5ex] \draw [color=black, thick, dashed] (0, 0) -- (0.5, 0); \end{tikzpicture} true AUROC, bounds based on 
\begin{tikzpicture}[baseline=-0.5ex] \draw [color=cyan!80!black, opacity=0.8] (0, -0.15) rectangle (0.7, 0.25); 
\fill [color=cyan!80!black, opacity=0.4] (0, -0.15) rectangle (0.7, 0.25); \end{tikzpicture} known positives and 
\begin{tikzpicture}[baseline=-0.5ex] \draw [color=red!80!black, opacity=0.8] (0, -0.15) rectangle (0.7, 0.25); 
\fill [color=red!80!black, opacity=0.4] (0, -0.15) rectangle (0.7, 0.25); \end{tikzpicture} known negatives.}
\label{parameffects-both}
\end{table}


\end{document}